# Decoupled Self Attention for Accurate One Stage Object Detection

Kehe WU[a], Zuge Chen[a*], Qi MA[b], Xiaoliang Zhang[a], Wei Li[a]

[a]School of Control and Computer Engineering, North China Electric Power University, Beijing, 102206, China
[b]China Mobile (Hangzhou) Information Technology Co., Ltd., Hangzhou, 311121, China

**Abstract:** As the scale of object detection dataset is smaller than that of image recognition dataset ImageNet, transfer learning has become a basic training method for deep learning object detection models, which will pretrain the backbone network of object detection model on ImageNet dataset to extract features for classification and localization subtasks. However, the classification task focuses on the salient region features of object, while the location task focuses on the edge features of object, so there is certain deviation between the features extracted by pretrained backbone network and the features used for localization task. In order to solve this problem, a decoupled self attention(DSA) module is proposed for one stage object detection models in this paper. DSA includes two decoupled self-attention branches, so it can extract appropriate features for different tasks. It is located between FPN and head networks of subtasks, so it is used to extract global features based on FPN fused features for different tasks independently. Although the network of DSA module is simple, but it can effectively improve the performance of object detection, also it can be easily embedded in many detection models. Our experiments are based on the representative one-stage detection model RetinaNet. In COCO dataset, when ResNet50 and ResNet101 are used as backbone networks, the detection performances can be increased by 0.4% AP and 0.5% AP respectively. When DSA module and object confidence task are applied in RetinaNet together, the detection performances based on ResNet50 and ResNet101 can be increased by 1.0% AP and 1.4% AP respectively. The experiment results show the effectiveness of DSA module. Code is at: https://github.com/chenzuge1/DSANet.git.

# 1 Introduction

ImageNet[1] dataset is a large scale image classification data set built by Professor Li Fei-Fei, which contains tens of millions of images and tens of thousands of categories of objects. It greatly promotes the development of image recognition task. Also it can improve other more complex Computer Vision(CV) tasks by transfer learning, such as object detection and instance segmentation. Object detection task contains classification and localization subtasks, and they focus on different spatial features of objects, such as classification task more focus on salient area features of object and localization task more focus on edge features of objects[2], so the features extracted by the pretrained backbone network are not entirely suitable for localization task. In order to learn suitable features for both tasks, some detectors try to be trained on object detection datasets from scratch[3-5]. Though their performances is close to detectors using transfer learning strategy, but they need more training time and more abundant expert experience. Therefore, most object detectors still prefer transfer learning in training at present. In order to use transfer learning in object detection, we proposes a method to extract features from pretrained features by introducing self attention mechanism, and it can automatically extract relevant features for specific task.

The attention mechanism comes from brain imaging mechanism research, when people observe a scene, they generally pay more attention to salient objects and pay less attention to background. Then it is introduced into deep learning networks to distinguish the importance of different features to task. Generally, weight is used to represent the importance of feature, the larger the weight, the more important the feature. In 2014, Bahdanau et al. firstly introduced attention into NLP machine translation task[6]. The translation model includes two processes: encoder and decoder, and they applied the attention mechanism to decoder stage to solve the problem of long distance dependence in machine translation. "Attention is all you need"[7] published by Google team in 2017 proposes a new machine translation model Transformer, which only uses attention mechanism instead of CNN or RNN to extract features in encoder stages, it greatly improves the performance of machine translation task. Then Google proposed a pretrained model Bert[8] in NLP field based on Transformer. Bert can not only reduce the training time of most NLP tasks, such as named entity recognition, machine translation and reading comprehension, but also greatly improve the performances of these tasks. The most important contribution of Transformer is that it proposes the feature extracted method using self attention mechanism, which is used to learn the relationship between self(input) and self(input), so it is called self-attention. Self attention can capture the internal correlation of all features, so each input feature can contain the whole input information. Recently, there are also some research works introducing attention mechanism into CV field. SENet[9] achieves best result in 2017 ImageNet competition, and its most important innovation is that it designs an attention module in channel domain. The module can learn the weight of each channel, and it represents the importance of each channel to output and it is used to recalibrate features for task, which makes output pay more attention to important channel features and suppress invalid channel features. Although the structure of channel attention of SENet is simple, the idea promotes the development of attention mechanism in CV field. CBAM[10] further explores the application scope of attention mechanism and proposes the combination of channel attention module and spatial attention module. The channel attention module is used to learn the channel features needed to be focused on, and the spatial attention module is used to learn the spatial location features needed to be paid attention to. Non-local Net[11] entirely uses self-] attention mechanism to learn features, which is very similar to Transformer. It enables each feature can learn global context information, so it solves the problem of small receptive field for features extracted by convolution network. This work can achieve excellent performances in object detection and video tracking tasks. However, there is too large amount of computation in the model. DANet[12] applies attention mechanism to image segmentation task. The smaller receptive field of convolution features results in the pixel from same object having different classes, which restricts the segmentation performance. It uses spatial and channel self attention module to extract features containing global spatial and channel information, which effectively improves segmentation performance.

From above we can see that attention mechanism has been applied to many image tasks, but there is few works on object detection. Object detection consists of two subtasks focusing on different spatial features of objects, and the attention mechanism can learn the importance of different spatial features. So it is natural to introduce attention into object detection. Besides, self attention is generally used in features extracted stage. Therefore, we propose a object detection model DSANet based on self attention mechanism, and it uses the spatial domain self attention module DSA (decoupled self attention) to extract suitable features for specific tasks, also DSA supplies two different features extracted branches for classification and localization tasks, that's why it is called decoupled self attention. In summary, the main contributions of this paper are described as follows:

1) To make full use of pretrained backbone network in one-stage object detection, we propose an decoupled module to extract suitable features for specific tasks.

2) As the two subtasks of object detection focus on different spatial features of objects, the decoupled module uses spatial self attention mechanism to learn more suitable features for different tasks, and the module is called as DSA in this paper.

3) In order to validate DSA module, we propose an object detection model DSANet based on

RetinaNet. All experiments are built on COCO dataset, DSA module can improve 0.4% AP and 0.5% AP with ResNet-50 and ResNet-101[13] backbone networks respectively. We further build an object detection model DSANet-Conf, which applies both DSA module and object confidence subtask[14] to RetinaNet. It can gain 36.3% AP and 38.4% AP with ResNet-50 and ResNet-101 backbone networks respectively. The experiment results shows that DSA module can improve one-stage object detection model.

# 2 Related Work

To extract suitable features for two tasks in object detection model, many works have been published. At present, these works are mainly divided into three categories: (1) improvement on backbone network; (2) training from scratch; (3) improvement on head networks of subtasks.

**(1) Improvement on backbone network**. This strategy is mainly used to improve the semantic information and location information of features at the same time. At present, the most commonly used backbone network for object detection is FPN[15], which adds upper sampling operations to the original image classification model and fuses the top and down feature maps. The fused features would include better semantic information and location information, so it can be shared by classification and localization tasks. PANet[16] thinks that the location information of top-level features in FPN has been greatly weakened by dozens or hundreds of convolution layers. Therefore, PANet proposes to add a bottom-up convolution pathway on the basis of FPN, so that the location information of top-level features can be enhanced. In order to improve detection performance without significantly increasing model parameters and computation, a multiscale object detection method STDN[17] based on DenseNet[18] is proposed. As each layer features of DenseNet contains the information of all previous layers, so STDN only uses the last feature map to detect objects. Pooling and special designed up-sampling method are used on the last feature map to generate different scale feature maps, and they would be used to detect different scale objects. M2Det[19] uses TUMs to refine features, which can increase model depth to improve context information, and the combination of TUM and FFM makes features still contain reliable location information with the deepening of model. M2Det gathers the both advantages of FPN and PANet, so it is more suitable for object detection. According to G-FRNet[20] proposes a gating unit to filter the fuzzy semantics in low-level features through high-level features, which enables the low-level features have strong ability to distinguish object categories under fine-grained location information. NAS-FPN[21] uses neural architecture search (NAS) to automatically search an optimal FPN NAS-FPN, which can better balance model accuracy and efficiency. EfficientDet[22] proposes an efficiency and accurate object detection model based on PANet after analyzing the NAS-FPN and PANet, and it can gain same excellent performance with PANet when the number of model parameters is greatly reduced. The above improved works are basically based on FPN, which use feature fusion strategy to improve the context information or location information of features. Different from them, RFBNet[23] improves the semantic information of low-level features by introducing dilated convolution. It uses dilated convolution with different dilated rates to generate features with different receptive fields, and these features will be concatenated to detect objects. RFBNet applies this strategy to the low-level features of SSD to improve the context information, so as to improve the detection performance of small objects.

**(2) Training from scratch**. "Rethinking ImageNet Pre-training"[3] published by Kaiming He et al. proves that object detection without pretrained backbone network can still converge after more training iterations. However, DSOD[5] finds that it is instability to train detection model from scratch. Then ScratchDet[4] delves into this problem and it finds that the main reason is that BN is not used in the model. When BN is added to model, it can train stably and its performance is better than that of pretrained model. DetNet[24] is a specific backbone network designed for object detection task. It uses dilated convolution to increase the receptive field of features without reducing the scale of feature map. So it can not only retain the location details of objects, but also improve the context information of features.

(3) **Improvement on head networks**. These improved research works are used to design a special feature extraction network for each subtask, and the network is generally located between backbone network and subtasks, which is generally called head network. RetianNet[25] firstly proposed head network, which includes four convolution layers for every subtask, so it can extract specific features for each task. The detection performance will be reduced by about 10% AP without head networks, which shows the importance of extracting independent features for each subtask. Double head RCNN[26] finds that the feature extracted by full connected network is more suitable for classification, while the feature extracted by convolution network is more suitable for localization. So it adds different networks after RoI(Region of Interest)[27] pooling features to extract features for different tasks. As the head networks of the two subtasks are different, the improved model is called Double Head RCNN. Inspired by Double Head RCNN, Senstime team delves into the differences of the features used for two subtasks[2]. They finds that the salient region features and edge features of objects have the greatest impact on classification and localization tasks respectively. In hence, they proposed a TSD module to extract suitable features for different tasks from ROI pooling features. And TSD uses different style deformable convolutions to extract different spatial features for different tasks, so as to decouple the spatial features of localization and classification tasks.

According to above analysis, the "improvement on backbone network" method needs more expert experiences to design network architecture. Although it can improve the feature presentation, it needs more time and computation to find the optimal architecture from infinite solution space. Although the "training from scratch" method can learn the suitable features for different tasks, it needs to configure parameters carefully and needs more training time. The "improvement on head networks" method does not modify the backbone network, and it only adds an independent head network for each task, so we can still use pretrained network, also it can extract individual features for different tasks, so it can extract appropriate features with less training time. Therefore, we improve the performance of detection model based on the "improvement on head networks" method. In order to reduce human experience in head network design, the head network in this paper uses self attention mechanism to automatically extract features for each task.

# 3 Methods

In this section we will introduce DSANet detailly. Firstly we show the network architecture of DSANet, which adds an DSA module to RetinaNet. Then we introduce the attention mechanism used in DSA and we analyze the difference of different attention operations. Finally we introduce the training and inference processes of DSANet simply.

## 3.1 Architecture of RetinaNet-Conf Detector

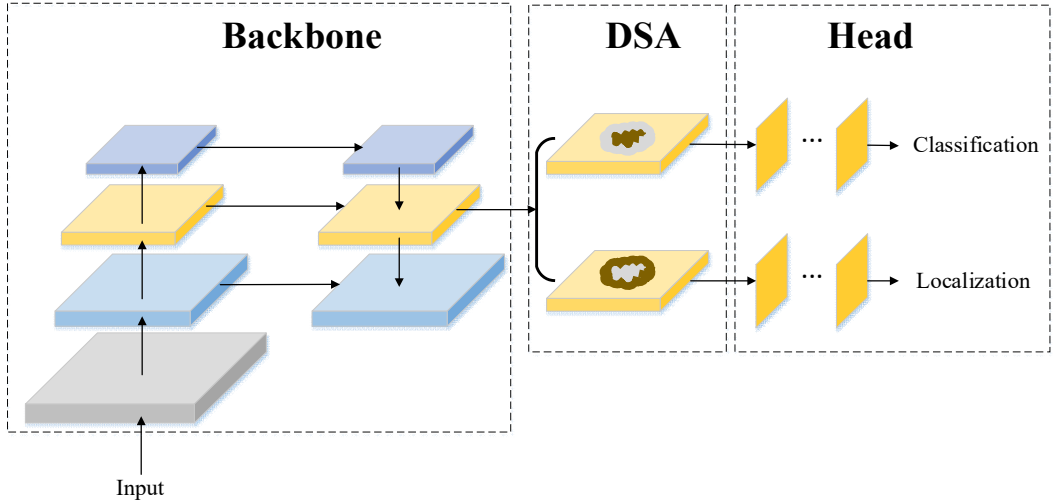

Figure 1 The network architecture of DSANet object detection model. DSANet is built based on RetinaNet, and it includes three modules: backbone network module, DSA module and head network module. Compared with RetinaNet, DSANet adds an DSA module between backbone network and head network. DSA uses self attention mechanism to learn suitable features for tasks, also it includes two branches, and each branch is unique to each task.

DSA module consists of two branches using self attention mechanism to extract features for classification and localization tasks. As shown in Figure 1, the top branch of DSA extracts the salient region features of objects for classification task, and its down branch extracts the edge features of objects for localization task. Existed object detection models generally use FPN as backbone network, then use its different scales features to detect different scales objects. DSANet adds a DSA module after each fusion features of FPN to extracted attentioned features of different scales objects. Besides, DSA only uses self attention mechanism in spatial domain of convolution features, because classification task and location task focus on different spatial features of objects.

DSA can be calculated by function(1):

$$Feature\_att = Attention(f(Feature\_input)) \quad (1)$$

In funciton(1), $Feature\_input$ is the input of DSA, $Feature\_att$ is the output of attention operation, $f$ represents preprocess functions on input, and it generally includes max-pooling, average-pooling and 1*1 convoluton operations. $Attention$ represents the attention operation, it can use learned method or vector similarity calculation method to define the weights of features. In this paper, $Attention$ is self attention, so $f$ will use three 1*1 convolution network to generate three features representing input, and the similarity method is used to calculate weights of spatial features through the correlation between them.

Then we can use function(2) to calculate the output of DSA module:

$$Feature\_out = Feature\_input + gamma * Feature\_att \quad (2)$$

In function(2), $Feature\_out$ represents the output of DSA module. From the function we can see that DSA module uses the same calculation method as residual module of ResNet, and its input is directly connected to output. The attention output features works as a branch of DSA output, and $gamma$ is a learned parameter, which is used to banlance the improtance of initial input and attention output. So DSA module with residual design can focus on important feature without increasing the model training difficulty.

As self attention is used in DSA module, so each feature of its output and attention output contains global spatial context information. DSA module is located before head network, which ensures each feature of head network will contain global spatial context information, which is critial to classifcation and localization tasks, so DSA module can improve object detection performance, especially for small objects.

From above description we can know that DSA use saptial attention to extract features. Now there are two spatial attention methods in computer vision: 1) Using convolution network to learn attention weight for each spatial postion of objects, and the spatial attention feature map is the product of attention weight and input features; 2) Using self attention mechanism to calculate the weights between either trwo spatial position features, and each spatial position feature of output is the weighted sum between weights and values of other spatial position features. Figure 2 shows the two spatial attention feature maps generated methods.

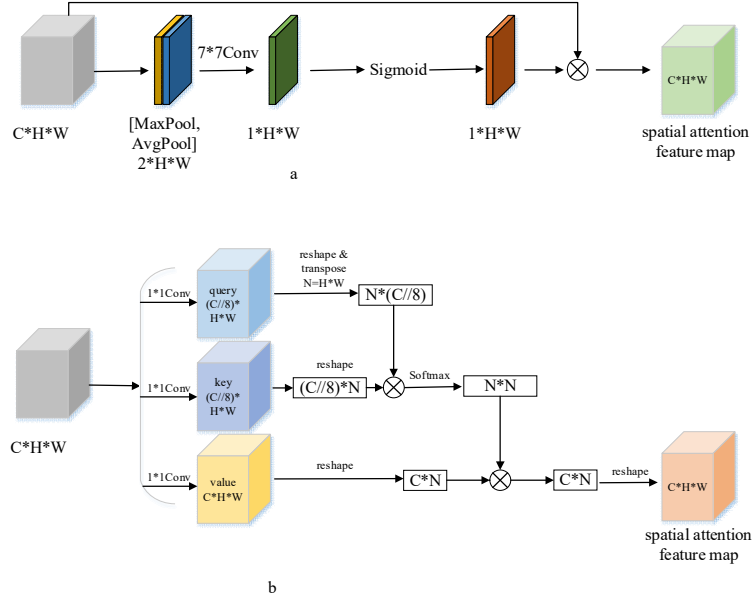

Figure 2 Two spatial attention feature maps generated methods. (a) shows the spatial attention method used in CBAM, and (b) shows the spatial attention methods used in DSANet. In our work, DSA module uses attention method shown in figure 2(b) to extract features.

The spatial attention method in Figure 2(a) is inspired by channel attention of SENet. The channel attention of SENet firstly calculates the mean of each channel features, and the mean value works as the input of two-layer fully connected(fc) network, and the output of fc network is the weight of each channel. Figure 2(a) need calculate the maximum and mean of each spatial position features, then a 7*7 convolution are used to reduce feature dimension from 2-D to 1-D. The 1-D fetures will be used as the input of sigmoid function, and the output is the weight of each spatial position feature, and all channels share the weight of same spatial position. The spatial attention features can be calculated by function(3) and function(4):

$$\begin{aligned} M_S(F) &= \sigma(Conv^{7*7}([MaxPool(F), AvgPool(F)])) \\ &= \sigma(Conv^{7*7}([F_{max}^S, F_{avg}^S])) \end{aligned} \quad (3)$$

$$F_{spatial\_att} = F \otimes M_S(F) \quad (4)$$

In function(3), $M_S(F)$ represents the weight results of all spatial position of objects, $F$ is the input of attention module. $MaxPool$ and $AvgPool$ represents max pooling and average pooling operation, and $F_{max}^S$ and $F_{avg}^S$ are the output features after different pooling operations. Function(4) shows the calculation of spatial attention features through the input and the weights of spatial position.

Figure2(b) shows the spatial attention features calculation process based on self attention, which is introduced from the machine translation model Transformer, also Vaswani et al. summarized the attention function[7], which is shown as function(5):

$$Attention(Q,K,V) = soft\max(\frac{QK^T}{\sqrt{d_k}})V \qquad (5)$$

In function(5), $Q$, $K$, $V$ represent Query, Key and Value. Query is a query matrix, used to represent the affected samples. If the dimension of query matrix is $N*d_k$, and $N$ represents the number of affected samples, $d_k$ represents the dimension of features in each samples. Key and Value are the different representations of impact samples, and [Key, Value] can be seen as the key/value pair of impact samples, and Key is used to identify which sample, while Value represents the value of the sample, so they are one-to-one match. If the dimensions of them are $M*d_k$, $M$ represents the number of impact samples and $d_k$ represents the dimension of features in impact sample. We can find the feature dimensions of impact sample and affected samples are same. $QK^T$ is the matrix product of matrix $Q$ and $K^T$, which shows the similarity results of affected samples and impact samples. $soft\max(\frac{QK^T}{\sqrt{d_k}})$ can be used to calculate the weights between all impact samples and each affected sample, so the product of it and matrix Value is the spatial attention fetures. Function(5) shows the common attention calculation function, the matrix Query, Key and Value are all be representations of impact samples when it represents self attention. As the source of Query matrix is same with that of Key and Value, so this attention ia called as self attention.

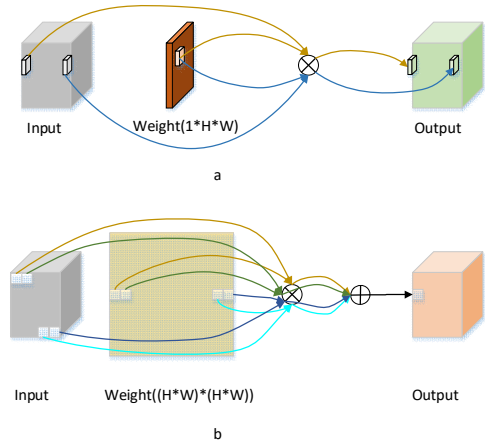

Figure 3 The different correlations between output and input based on two different spatial attention methods. (a) shows the correlations based on figure 2(a), and (b) shows the correlations based on figure 2(b). The lines with same color represent a pair of input and weight.

From Figure 3(a) we can see that each spatial feature of output is only related to one spatial feature of input, so the output features can't contain the global spatial context information. However, from figure 3(b) we can see that each spatial feature of output is related to every spatial feature of input, so every feature of output contains global spatial context information. In addition, each spatial feature of output owns the unique weights between other spatial features with this feature, so the focused area of each spatial feature in output is different. In summary, the attention in Figure 3(a) is essentially different from the attention in Figure 3(b). Figure 3(a) uses convolution network to learn the weights of each spatial feature from channel global information. The weight represent the importance of each spatial feature in next layer features. Therefore, the receptive field of output feature is same with that of input feature, which shows the semantic information of feature is not improved with the deep increasing of network. Besides, the filter size of convolution is 7*7, so it will bring large amount of parameters and computation, which will increase model training difficulty. However, figure 3(b) using self attention can not only use global receptive field information to improve feature representation, but also not bring many additional model parameters. But we should note that self attention needs more computation cost, because the computation of $soft\max(\frac{QK^T}{\sqrt{d_k}})V$ is $d_k*N*N*N$, $N$ is equal to $H*W$, and $d_k$ is the number of channels. When using self attention mechanism on the feature map with higher resolution, the huge computation will result in out of GPU memory. Therefore, it is need to carefully select the feature map adding DSA module when computation resources is insufficient. However, from the characteristics of CNN we can know that the receptive field of feature map with higher resolution is generally smaller, and they represents less context information, so these feature map more need DSA module, but they nedd more computation. Therefore, it is a worthy research task how to balance the context information and computation of high-resolution feature map with DSA module.

### 3.2 Training & Inference

DSANet adds an DSA module in the mid of RetinaNet. And model training process mainly updates network parameters according to the loss function. As the DSANet modify model task, so it has same loss function with RetinaNet. So the loss function of DSANet also includes two parts: classification loss function and localization loss function. We will not repeat it here again. As DSA module is added after the FPN network, wo we can still use pretrained ResNet parameters. Therefore, the added DSA module will not slow the parameters training efficiency of FPN and head network. Therefore, the initialization and training setting of these parameters can be consistent with RetinaNet. From function(2) we can see a new parameter $gamma$ is added to measure the percentage of attention feature map in output of DSA. In order to train this parameter, we initialize it to 0 and update it using the same update strategy as other network parameters in the training process. The inference process of model mainly focuses on the selection strategy of box prediction results, such NMS. Similarly, the DSA module will not affect the selection process of detection results, so the inference process is also consistent with RetinaNet.

# 4 Experiments

In order to validate DSANet, we use the large scale benchmark object detection dataset MS coco 2017 in experiments. Train2017 is used for training and Val2017 is used for validation, and all APs and Recalls indexes defined by coco dataset are used for evaluation. As DSANet is based RetinaNet and MMDetection[28] is an excellent open CV task sources platform by Shang Tang, so all experiments in this paper are performed on the RetinaNet source code in MMDetection. The experiment environment includes 16G memory, 8-core CPU, and a Tesla V100 GPU with 32gG memory. The input size of the model is [1000, 600], batch size is 16 and total training epoch is 12. The optimizer is SGD with initial learning rate 0.01, also the learning rate will decay 0.1 times in epoch [9, 12].

| Model_Name | AP | $AP_{50}$ | $AP_{75}$ | $AP_S$ | $AP_M$ | $AP_L$ | AR | $AR_S$ | $AR_M$ | $AR_L$ |
|---|---|---|---|---|---|---|---|---|---|---|
| RetinaNet | 0.353 | 0.543 | 0.375 | 0.184 | 0.394 | 0.480 | 0.522 | 0.308 | 0.573 | 0.694 |
| DSANet_a(3-7) | 0.356 | 0.545 | 0.380 | 0.184 | 0.393 | 0.487 | 0.525 | 0.306 | 0.575 | 0.701 |
| DSANet_b(4-7) | 0.357 | 0.548 | 0.379 | 0.185 | 0.399 | 0.483 | 0.524 | 0.308 | 0.575 | 0.693 |

Table 1 Comparison of DSA module using different attention methods. DSANet_a uses the attention method shown in figure 2(a), while DSANet_a uses the attention method shown in figure 2(b). "3-7" or "4-7" shows that we add DSA module in FPN conv3-7 or conv4-7 fusion features. As the computation of self attention is too huge, so it is out of GPU memory when using self

attention in conv3 feature map with higher resolution. In hence, DSANet only adds self attention DSA module in conv4-7 FPN fusion features, and other experiments abide by the same agreement.

From table 1 we can see that, DSANet_a and DSANet_b all achieve better detection performance than RetinaNet, which shows DSA module with either attention methods can improve object detection task. Although DSANet_b does not add DSA in conv3 feature map, DSANet_b still performs better than DSANet_a, which shows the features extracted by self attention have better presentation, as it contains global context information. Compared with DSANet_a, AP50 and APM of DSANet_b are improved by 0.3%AP and 0.6%AP, but APL and ARL of it has been reduced by 0.4%AP and 0.8%AP, and other indexes are nearly same in two models. The improvement of AP50 shows that DSANet_b can improve detection performance of hard examples. Although DSA module is not added in conv3 feature map, which is used to detect small objects, the values of APs, APM, ARS and ARM have been improved, while APL and ARL are reduced, the reason for this phenomenon may be that because there is no presentation improvement of conv3 features, which results in the losses of smaller objects are larger, so these losses will be the main factor guiding model training, in hence, the performances of smaller objects are improved while indexes of larger objects are reduced. And it is a important subject needed to delve into continually. As the resolution of conv3 feature map is larger, the number of anchor boxes based on it is more than the sum of other all anchor boxes, also DSANet_b(4-7) has improved performance, so it is worthy to believe that it can improve detection performance further when conv3 feature map adding DSA module, but as the restrict of GPU computation, it is hard to validate the conclusion. So if no otherwise specified, DSANet in later experiments all represent DSANet_b(4-7).

| Model_Name | AP | $AP_{50}$ | $AP_{75}$ | $AP_S$ | $AP_M$ | $AP_L$ | AR | $AR_S$ | $AR_M$ | $AR_L$ |
|---|---|---|---|---|---|---|---|---|---|---|
| RetinaNet | 0.353 | 0.543 | 0.375 | 0.184 | 0.394 | 0.480 | 0.522 | 0.308 | 0.573 | 0.694 |
| DSANet(share) | 0.356 | 0.547 | 0.377 | 0.187 | 0.397 | 0.474 | 0.524 | 0.309 | 0.576 | 0.694 |
| DSANet | 0.357 | 0.548 | 0.379 | 0.185 | 0.399 | 0.483 | 0.524 | 0.308 | 0.575 | 0.693 |

Table 2 Comparison of whether classification and localization tasks share DSA module or not. DSANet(share) represents that classification and localization tasks of DSANet share DSA module, namely, DSA only has one branch. DSANet shows that DSA provides two different branches for classification and localization tasks. And DSAnet(share) and DSANet all add DSA module in conv4-7 feature maps.

In order to the features extracted by pretrained backbone network are more suitable for classification and localization tasks, RetinaNet adds head network with two branches to refine features for different tasks. In order to verify whether the DSA refinement module is universal to two subtasks, we design experiments shown in table 2. Compared to base RetinaNet, DSANet(share) and DSANet both improve performance, while DSANet gains 0.1% AP than DSANet(share), which show that DSA module can improve feature representation indeed. Though DSANet performs slight better than DSANet(share) in AP index, but for other indexes, except for APS, all other indexes have been improved, also APL is greatly increased by 0.9%, which shows the false detection rate is reduced, especially for medium and large objects. For all recall indexes, their total AR are same, but other indexes of DSANet are slight lower than those of DSANet(share), which shows the missed detection rate of DSANet is increased. So decoupled self attention can extract more suitable features for different tasks from backbone network features, so it can gain higher precision, and almost all AP indexes have been effectively improved. But the missed detection rates of it are increased in all scales objects, it may be caused by dense objects, because the detection performance of these objects has been improved, but they are too dense to be left in NMS. Therefore, share of DSA module can be regarded as a strategy to balance false and missed detection rate requirements. If a task more concern false detection rate, it can choose DSANet and it will need more computation, while a task concern missed detection rate, it can choose DSANet(share).

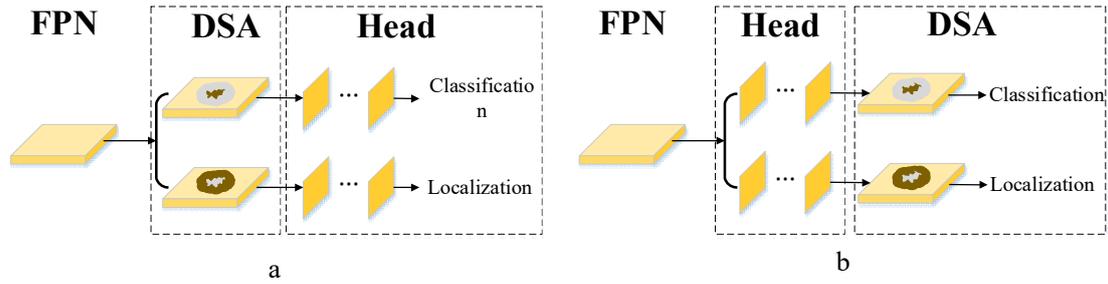

Figure 4 Different location of DSA module in RetinaNet. Figure 4(a) locates DSA module between FPN and Head network, while figure 4(b) locates DSA module after Head network.

| Model_Name | AP | $AP_{50}$ | $AP_{75}$ | $AP_S$ | $AP_M$ | $AP_L$ | AR | $AR_S$ | $AR_M$ | $AR_L$ |
|---|---|---|---|---|---|---|---|---|---|---|
| RetinaNet | 0.353 | 0.543 | 0.375 | 0.184 | 0.394 | 0.480 | 0.522 | 0.308 | 0.573 | 0.694 |
| DSANet(after) | 0.355 | 0.545 | 0.377 | 0.183 | 0.399 | 0.483 | 0.524 | 0.307 | 0.574 | 0.689 |
| DSANet(before) | 0.357 | 0.548 | 0.379 | 0.185 | 0.399 | 0.483 | 0.524 | 0.308 | 0.575 | 0.693 |

Table 3 Comparison of different DSA module location. DSANet(befor) represents the location of DSA module is between backbone and head network, and DSANet(after) represents the location of DSA module is after head network.

In order to place DSA module in right location of RetinaNet, the experiments shown in table 3 are designed. In order to use pretrained backbone network to train efficiently, we select two locations of DSA module shown in figure 4. As shown in table 3, DSANet(before) performs better than DSANet(after), and it not only increases by 0.2% in AP, but also improves all other evaluation indexes comprehensively. It shows that no matter where the DSA module is located in the RetinaNet, although their performance is different, their improvement direction is same, namely, DSANet(before) performs better than DSANet(after) in all of indexes. Besides, when DSA module is located before head network, as it can extract features containing global context information from FPN fusion features, so all of head features can also contain global context information, so the features will have stronger presentation. While DSA module is located after head network, only the features of DSA module contain global information, which will reduce feature representation. Therefore, it is necessary to locate DSA module in lower feature layer as far as possible, so that more feature layers can learn from global context to improve features, so as to improve detection performance.

| Model_Name | AP | $AP_{50}$ | $AP_{75}$ | $AP_S$ | $AP_M$ | $AP_L$ | AR | $AR_S$ | $AR_M$ | $AR_L$ |
|---|---|---|---|---|---|---|---|---|---|---|
| RetinaNet | 0.353 | 0.543 | 0.375 | 0.184 | 0.394 | 0.480 | 0.522 | 0.308 | 0.573 | 0.694 |
| DSANet(gamma=1) | 0.355 | 0.548 | 0.376 | 0.184 | 0.400 | 0.480 | 0.524 | 0.304 | 0.579 | 0.694 |
| DSANet | 0.357 | 0.548 | 0.379 | 0.185 | 0.399 | 0.483 | 0.524 | 0.308 | 0.575 | 0.693 |

Table 4 Comparison of DSA module with different gamma setting. DSANet(gamma=1) represents the value of parameter gamma in function(2) is set to 1.

From function(2) we can know that there is a learned parameter gamma used to define the weight of spatial attention feature in DSA output. We design the experiments shown in table 4 to evaluate the necessary to learn gamma. From table 4 we can see that when gamma is set to 1, DSANet can increase by 0.2% AP than RetinaNet, while DSANet with learned gamma can increase by 0.4% AP, so learned gamma performs better than constant gamma. The phenomenon shows that DSA with learned gamma can extract more suitable and flexible features for different tasks. The DSANet has same AP50 with DSANet(gamma=1), while it has better AP75, which shows the learned gamma is critical to localization task, so the index with bigger IoU threshold performs better. For different scales objects, when learning gamma in training process, the APS, ARS and APL increase by 0.1%, 0.4% and 0.3%, while the ARL, APM and ARM reduce by 0.1%, 0.1% and 0.4%, which shows that the constant gamma is more suitable for medium scale object detection task. It indicates that input feature and spatial attention feature are same important to output feature of DSA module for objects with specific scales, while they play different roles for large or small objects. Also it is hard to learn the value 1 for gamma of medium objects, so their detection performance is lower in DSANet with learned gamma. In hence, there are different gamma values for different scales objects, also DSANet achieves better AP than DSANet(gamma=1), so we will learn gamma values in model training in next experiments.

As described in section 3.1, the computation of DSA based on conv3 feature map will be out of GPU memory, we find the $soft\max(\frac{QK^T}{\sqrt{d_k}})V$ calculation is the main reason, its computation is C*(W*H)*(W*H)*(W*H). Because width and height of conv3 feature map is large, so they bring amount of computation. In order to apply DSA to conv3 feature map, we should decrease W and H. And we try to use convolution to reduce the dimension of Query, Key and Value, which is shown in figure 5.

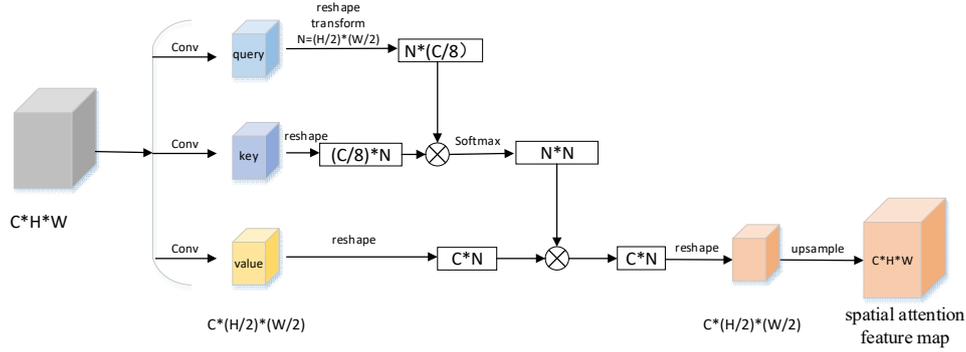

Figure 5 The computation process of DSA module used in conv3 feature map. Compared with figure 2 (b), the convolution used to generate query, key and value feature maps is different, it is used to reduce the spatial dimensions of them, so as to reduce the computation of DSA. But the dimension of DSA output will be reduced, so we next use nearest neighbor interpolation to recover the dimension of output.

The stride of convolution in figure 5 is set to 2, and the kernel size of it is 1*1 or 3*3, so the width and height of three transformed feature maps are half of input. Also in order to compare FPN and DSA, we also locate DSA in different scales feature maps of RetinaNet.

| Model | Kernel_size | AP | $AP_{50}$ | $AP_{75}$ | $AP_S$ | $AP_M$ | $AP_L$ | AR | $AR_S$ | $AR_M$ | $AR_L$ |
|---|---|---|---|---|---|---|---|---|---|---|---|
| RetinaNet | - | 0.353 | 0.543 | 0.375 | 0.184 | 0.394 | 0.480 | 0.522 | 0.308 | 0.573 | 0.694 |
| DSANet(4-7) | - | 0.357 | 0.548 | 0.379 | 0.185 | 0.399 | 0.483 | 0.524 | 0.308 | 0.575 | 0.693 |
| DSANet(3-7) | 3*3 | 0.355 | 0.545 | 0.376 | 0.181 | 0.395 | 0.481 | 0.522 | 0.299 | 0.573 | 0.694 |
| DSANet*(3-7) | 3*3 | 0.327 | 0.493 | 0.349 | 0.131 | 0.367 | 0.477 | 0.504 | 0.250 | 0.561 | 0.702 |
| DSANet(3-7) | 1*1 | 0.355 | 0.547 | 0.377 | 0.188 | 0.395 | 0.484 | 0.525 | 0.313 | 0.574 | 0.697 |
| DSANet*(3-7) | 1*1 | 0.328 | 0.493 | 0.352 | 0.135 | 0.370 | 0.379 | 0.501 | 0.248 | 0.562 | 0.692 |

Table 5 Comparison of DSA with different convolution used in conv3. "3-7" or "4-7" shows that DSA module added in FPN or ResNet conv3-7 or conv4-7 feature maps. DSANet* represents it uses ResNet instead of FPN as backbone network. Besides, the convolution with 3*3 kernel size need padding.

From table 5 we can see there is similar performance for DSA module used in conv3 feature map with different kernel sizes when using same backbone network, it shows that the kernel size of convolution is the main influence on detection performance. However, we can find DSANet(4-7) performs better than both of DSANet(3-7) with different kernel size, though it has less computation and model parameters. It may be that the last up sampling brings misplaced spatial weights for features, so it reduces the detection performance. Besides，the performances of DSANet with FPN backbone are better than those of DSANet with ResNet backbone, also RetinaNet performs better than DSANet with ResNet backbone, these two comparison results show that FPN module is critical to object detection models, whether it uses self attention mechanism. But it is hard to say whether FPN module performs better than DSA module for object detection task or not. Because DSA module proposed by figure 5 can reduce detection performance, and using the DSA module shown in figure 2(b) in conv3 feature map may be benefit to detection performance. From above analyses we can see that the combination of DSA and FPN can achieve the best results, so they will be used in detection model together in following experiments.

RetinaNet_Conf[14] is one of our proposed works, it proposed an object confidence subtask to solve the misalignment of classification and localization tasks. DSA module uses decoupled self attention branches to extract features for different task, it can also eases the misalignment of two subtasks. So next the combination of DSA module and object confidence task will be used in RetinaNet, and the new object detection model is named as DSANet_Conf.

| Model | AP | $AP_{50}$ | $AP_{75}$ | $AP_S$ | $AP_M$ | $AP_L$ | AR | $AR_S$ | $AR_M$ | $AR_L$ |
|---|---|---|---|---|---|---|---|---|---|---|
| RetinaNet(50) | 0.353 | 0.543 | 0.375 | 0.184 | 0.394 | 0.480 | 0.522 | 0.308 | 0.573 | 0.694 |
| RetinaNet(101) | 0.370 | 0.562 | 0.394 | 0.187 | 0.417 | 0.502 | 0.532 | 0.303 | 0.584 | 0.708 |
| RetinaNet-Conf(50) | 0.357 | 0.529 | 0.377 | 0.176 | 0.402 | 0.485 | 0.526 | 0.292 | 0.580 | 0.709 |
| RetinaNet-Conf(50_x) | 0.360 | 0.529 | 0.384 | 0.184 | 0.406 | 0.491 | 0.536 | 0.316 | 0.592 | 0.714 |
| RetinaNet-Conf(101) | 0.375 | 0.550 | 0.399 | 0.183 | 0.427 | 0.517 | 0.539 | 0.302 | 0.596 | 0.730 |
| RetinaNet-Conf(101_x) | 0.380 | 0.552 | 0.407 | 0.191 | 0.431 | 0.524 | 0.549 | 0.319 | 0.609 | 0.734 |
| DSANet(50) | 0.357 | 0.548 | 0.379 | 0.185 | 0.399 | 0.483 | 0.524 | 0.308 | 0.575 | 0.693 |
| DSANet(101) | 0.375 | 0.566 | 0.401 | 0.191 | 0.425 | 0.512 | 0.540 | 0.316 | 0.595 | 0.716 |
| DSANet-Conf(50) | 0.359 | 0.537 | 0.382 | 0.184 | 0.403 | 0.493 | 0.528 | 0.301 | 0.583 | 0.708 |
| DSANet-Conf(50_x) | 0.363 | 0.536 | 0.386 | 0.191 | 0.407 | 0.497 | 0.537 | 0.315 | 0.593 | 0.714 |
| DSANet-Conf(101) | 0.379 | 0.556 | 0.403 | 0.189 | 0.426 | 0.527 | 0.541 | 0.303 | 0.601 | 0.734 |
| DSANet-Conf(101_x) | 0.384 | 0.557 | 0.410 | 0.198 | 0.430 | 0.532 | 0.551 | 0.323 | 0.612 | 0.739 |

Table 6 Comparison of DSA module and object confidence task. '50' or '101' in model names represent using ResNet50 or ResNet101 as backbone network. 'x' represents using the product of classification score and object confidence to guide NMS process.

From table 6 we can see that when only using classification score to guide NMS, the AP of DSANet-Conf is increased by 0.2% than DSANet and RetinaNet-Conf based on ResNet50, while it is increased by 0.4% than them based on ResNet101. We can find the APs of DSANet and RetinaNet-Conf are always same, no matter the backbone is ResNet50 or ResNet101, while other evaluation indexes are different in two models, such as DSANet performs better on indexes with smaller IoU threshold and smaller scales, and RetinaNet-Conf performs better on other indexes, which shows that DSA module and object confidence task have different influence on detection task, so the combination of them can integrate their advantages to improve performance further. Also we can find that DSANet-Conf gains more improvement with deeper of backbone network, it indicates that the combination of two proposed strategies performs better based on more reliable feature representation. Compared with DSANet and RetinaNet-Conf, DSANet-Conf always gains better on AP75, APL and ARM, it shows that either of two strategies can improve the results of these indexes, so these indexes can be improved by their combination further. But for other indexes, two strategies have opposite influence, so the performance of their combination is compromise.

Here we will analyze the experiments of using the product of classification score and object confidence to guide NMS. Compared DSANet, DSANet-Conf can respectively gain more 0.6% AP and 0.9% AP with ResNet50 and ResNet101. As described in RetinaNet-Conf work, when object confidence is joined to guide NMS, the AP50 is reduced, while other indexes are improved. Compared with RetinaNet-Conf, DSANet-Conf can respectively gain more 0.3% AP and 0.4% AP with ResNet50 and ResNet101, and the experiment phenomenon is similar to adding DSA module in

RetinaNet, so we won't go into much detail here.

In summary, either of object confidence task and DSA module can improve detection performance, moreover, their combination can improve performance further, which not only proves the validation of the two ideas, but also shows that they improve detection task from different aspects. Therefore, we can adopt the two strategies simultaneously in practical application scene, so as to achieve the optimal detection result.

| detector | backbone | schedule | AP | $AP_{0.50}$ | $AP_{0.75}$ | $AP_S$ | $AP_M$ | $AP_L$ |
|---|---|---|---|---|---|---|---|---|
| YOLOv2[29] | DarkNet-19 | - | 0.216 | 0.440 | 0.192 | 0.500 | 0.224 | 0.355 |
| YOLOv3[30] | DarkNet-53 | - | 0.330 | 0.579 | 0.344 | 0.183 | 0.354 | 0.419 |
| SSD300[28, 31] | VGG16 | 20e | 0.257 | 0.439 | 0.262 | 0.069 | 0.277 | 0.426 |
| SSD512[28, 31] | VGG16 | 20e | 0.293 | 0.492 | 0.308 | 0.118 | 0.341 | 0.447 |
| Faster R-CNN[28, 32] | ResNet-50-FPN | 1x | 0.366 | 0.585 | 0.392 | 0.207 | 0.405 | 0.479 |
| Faster R-CNN[28, 32] | ResNet-101-FPN | 1x | 0.388 | 0.605 | 0.423 | 0.233 | 0.431 | 0.503 |
| Mask R-CNN[28, 33] | ResNet-50-FPN | 1x | 0.374 | 0.589 | 0.404 | 0.217 | 0.410 | 0.491 |
| Mask R-CNN[28, 33] | ResNet-101-FPN | 1x | **0.399** | **0.615** | **0.436** | **0.239** | **0.440** | 0.518 |
| FCOS[28, 34] | ResNet-50-FPN | 1x | 0.367 | 0.558 | 0.392 | 0.210 | 0.407 | 0.484 |
| FCOS[28, 34] | ResNet-101-FPN | 1x | 0.391 | 0.585 | 0.418 | 0.220 | 0.435 | 0.511 |
| RetinaNet[14] | ResNet-50-FPN | 1x | 0.353 | 0.543 | 0.375 | 0.184 | 0.394 | 0.480 |
| RetinaNet[14] | ResNet-101-FPN | 1x | 0.370 | 0.562 | 0.394 | 0.187 | 0.417 | 0.502 |
| IoU-aware Net[14] | ResNet-50-FPN | 1x | 0.361 | 0.527 | 0.389 | 0.189 | 0.406 | 0.489 |
| IoU-aware Net[14] | ResNet-101-FPN | 1x | 0.379 | 0.548 | 0.409 | 0.196 | 0.428 | 0.525 |
| RetinaNet-Conf[14] | ResNet-50-FPN | 1x | 0.360 | 0.529 | 0.384 | 0.184 | 0.406 | 0.491 |
| RetinaNet-Conf[14] | ResNet-101-FPN | 1x | 0.380 | 0.552 | 0.407 | 0.191 | 0.431 | 0.524 |
| RetinaNet-Conf[14] | ResNet-101-FPN | 2x | 0.384 | 0.555 | 0.412 | 0.194 | 0.433 | **0.535** |
| DSANet | ResNet-50-FPN | 1x | 0.357 | 0.548 | 0.379 | 0.185 | 0.399 | 0.483 |
| DSANet | ResNet-101-FPN | 1x | 0.375 | 0.566 | 0.401 | 0.191 | 0.425 | 0.512 |
| DSANet-Conf | ResNet-50-FPN | 1x | 0.363 | 0.536 | 0.386 | 0.191 | 0.407 | 0.497 |
| DSANet-Conf | ResNet-101-FPN | 1x | 0.384 | 0.557 | 0.410 | 0.198 | 0.430 | 0.532 |

Table 7 Comparison of state-of-the-art detectors. The table is divided into four parts. The first part shows the performances of yolov2 and yolov3 from their published works. And the second part shows the performances of state-of-the-art detection models, which are reported by MMDetection work. The third part and forth part show performances of improved RetinaNet, and these experiment results reported by our work.

From table 7 can be seen that Mask R-CNN with ResNet-101-FPN still achieves the best AP result on COCO dataset, except for APL, it performs best on other all indexes. And we can find DSANet-Conf has achieved the best performance in one stage object detection. Compared with base model RetinaNet, it increases by 1.0% AP and 1.4% AP with ResNet50 and ResNet101 respectively. Also DSANet-Conf can achieve the same performance with RetinaNet-Conf using two times training epochs, it shows that DSA module can improve detection performance with less training time, so DSA module is benefit to detection task. Besides DSANet with ResNet101 performs 0.1% AP better than Mask R-CNN with ResNet50, and APM and APL of DSANet have been increased by 1.5% and 2.1%, while APS of DSANet has been reduced by 2.6%, which shows DSA module can improve detection performance of medium and large scale objects and reduce detection performance of small objects. That's because DSA module is not added to conv3 feature map, which is used to detect small objects, but the improvement of larger objects detection validates DSA module. Also DSA module can be added to conv3 feature map when the computation is sufficient, we believe the detection task performance can be improved further. FCOS in table 7 is an anchor free detection model, it doesn't set large amount of anchor boxes for training, while it use same backbone network as one stage detection models, also it includes classification and localization subtasks, so DSA module can be embedded into FCOS model easily, so we can apply DSA module to different detection models to validate it further.

# 5 Conclusion

As the classification and localization subtasks in object detection model focus on different spatial features of objects, so we propose a DSA module based on self attention mechanism to extract different spatial attention features for two subtasks. DSA module is located between FPN backbone and head networks, so it can not only improve training efficiency through using pretrained backbone networks, but also solve the problem of features extracted by pretrained model are not suitable for localization task. Then we design abundant experiments to validate DSANet based on COCO dataset. The experiment results show that DSANet can increase by 0.4% AP and 0.5% AP than RetinaNet with ResNet50 and ResNet101 respectively. Also we apply DSA module and object confidence task, which is proposed by our another work, to RetinaNet, and the new model is called as DSANet-Conf, it can achieve the best performance of one stage object detection models. Without whistles and bells, it can achieve 38.4% AP. Besides, this result is gained without adding DSA module to conv3 feature map of FPN, which is used to detect smaller objects and includes the most of anchor boxes, because the larger computation of self attention in conv3 feature map can cause the out of GPU memory. The receptive field of conv3 feature map is smaller, so it more needs to use DSA to extract global spatial context. Therefore, it is worthy to introduce self attention into feature maps with higher resolution in future works.